\def\cvprPaperID{10924}
\def\confName{CVPR}
\def\confYear{2023}

\def\authorBlock{
    \textbf{Hao Zhang$^{1,4}$\thanks{Equal contribution.}\hspace{.4mm} \thanks{This work was done when Hao Zhang and Feng Li were interns at IDEA. }\hspace{1.5mm}, ~Feng Li$^{1,4*\dag}$, ~Huaizhe Xu$^{1,4}$, ~Shijia Huang$^{2,4}$},\\ \textbf{~Shilong Liu$^{3,4}$, ~Lionel M. Ni$^{5,1}$, ~Lei Zhang$^{4}$\thanks{Corresponding author.}\hspace{1.mm}} \\
$^1$The Hong Kong University of Science and Technology. \\
$^2$The Chinese University of Hong Kong. \\
$^3$Dept. of CST., BNRist Center, Institute for AI, Tsinghua University. \\
$^4$International Digital Economy Academy (IDEA). \\
$^5$The Hong Kong University of Science and Technology (Guangzhou).\\
}

\newif\ifreview 
\newif\ifarxiv \newcommand{\arxiv}{\arxivtrue}
\newif\ifcamera 
\newif\ifrebuttal 

\arxiv
\pdfoutput=1
\documentclass[10pt,twocolumn,letterpaper]{article}
\ifreview \usepackage[review]{cvpr} \fi
\ifarxiv \usepackage[pagenumbers]{cvpr} \fi
\ifrebuttal \usepackage[rebuttal]{cvpr} \fi
\ifcamera \usepackage{cvpr} \fi

\usepackage{graphicx}
\usepackage{amsmath}
\usepackage{amssymb}
\usepackage{booktabs}

\usepackage{times}
\usepackage{microtype}
\usepackage{epsfig}
\usepackage[table,xcdraw]{xcolor}
\usepackage{caption}
\usepackage{float}
\usepackage{placeins}
\usepackage{color, colortbl}
\usepackage{stfloats}
\usepackage{enumitem}
\usepackage{tabularx}
\usepackage{xstring}
\usepackage{multirow}
\usepackage{xspace}
\usepackage{url}
\usepackage{subcaption}
\usepackage{xcolor}

\usepackage{graphicx}
\usepackage{amsmath}
\usepackage[utf8]{inputenc} 
\usepackage[T1]{fontenc}    
\usepackage{url}            
\usepackage{booktabs}       
\usepackage{amsfonts}       
\usepackage{nicefrac}       
\usepackage{microtype}      
\usepackage{xcolor}         
\usepackage{adjustbox}

\usepackage{bbding}
\usepackage{multirow}
\usepackage{booktabs}
\usepackage{makecell}
\usepackage{bbm}
\usepackage{courier}
\usepackage{subfloat}
\usepackage{algorithmic}
\usepackage{tablefootnote}
\usepackage{tabulary,multirow,overpic,xcolor,subfloat}

\usepackage[hang,flushmargin]{footmisc}

\ifcamera \usepackage[accsupp]{axessibility} \fi

% ...

\newcommand{\M}{MP-Former }

\newcommand{\app}{\raise.17ex\hbox{$\scriptstyle\sim$}}

\newcolumntype{x}[1]{>{\centering\arraybackslash}p{#1pt}}
\newcolumntype{y}[1]{>{\raggedright\arraybackslash}p{#1pt}}

\newlength\savewidth\newcommand\shline{\noalign{\global\savewidth\arrayrulewidth
  \global\arrayrulewidth 1pt}\hline\noalign{\global\arrayrulewidth\savewidth}}
\newcommand{\tablestyle}[2]{\setlength{\tabcolsep}{#1}\renewcommand{\arraystretch}{#2}\centering\footnotesize}

\ifarxiv  \fi

\newcommand{\R}[1]{{%
    \textbf{%
        \ifstrequal{#1}{1}{\textcolor{red}{R#1}}{%
        \ifstrequal{#1}{2}{\textcolor{blue}{R#1}}{%
        \ifstrequal{#1}{3}{\textcolor{magenta}{R#1}}{%
        \ifstrequal{#1}{4}{\textcolor{teal}{R#1}}{%
                           \textcolor{cyan}{R#1}%
        }}}}%
    }%
}}

\usepackage{xr-hyper}

\makeatletter
\newcommand*{\addFileDependency}[1]{
  \typeout{(#1)}
  \@addtofilelist{#1}
  \IfFileExists{#1}{}{\typeout{No file #1.}}
}

\makeatother

\usepackage[pagebackref,breaklinks,colorlinks]{hyperref}
\usepackage[capitalize]{cleveref}
\crefname{section}{Sec.}{Secs.}
\crefname{table}{Table}{Tables}
\crefname{figure}{Fig.}{Figs.}

\begin{document}
\title{MP-Former: Mask-Piloted Transformer for Image Segmentation}
\author{\authorBlock}
\maketitle
\begin{abstract}
We present a mask-piloted Transformer which improves masked-attention in Mask2Former for image segmentation. The improvement is based on our observation that Mask2Former suffers from inconsistent mask predictions between consecutive decoder layers, which leads to inconsistent optimization goals and low utilization of decoder queries. To address this problem, we propose a mask-piloted training approach, which additionally feeds noised ground-truth masks in masked-attention and trains the model to reconstruct the original ones. Compared with the predicted masks used in mask-attention, the ground-truth masks serve as a pilot and effectively alleviate the negative impact of inaccurate mask predictions in Mask2Former. Based on this technique, our \M achieves a remarkable performance improvement on all three image segmentation tasks (instance, panoptic, and semantic), yielding $+2.3$AP and $+1.6$mIoU on the Cityscapes instance and semantic segmentation tasks with a ResNet-50 backbone. Our method also significantly speeds up the training, outperforming Mask2Former with half of the number of training epochs on ADE20K with both a ResNet-50 and a Swin-L backbones. Moreover, our method only introduces little computation during training and no extra computation during inference. Our code will be released at \url{https://github.com/IDEA-Research/MP-Former}.

\end{abstract}

\section{Introduction}
Image segmentation is a fundamental problem in computer vision which includes semantic, instance, and panoptic segmentation. Early works design specialized architectures for different tasks, such as mask prediction models~\cite{he2017mask,chen2019hybrid} for instance segmentation and per-pixel prediction models~\cite{long2015fully,ronneberger2015u} for semantic segmentation. Panoptic segmentation~\cite{kirillov2019panoptic} is a unified task of instance and semantic segmentation. But universal models for panoptic segmentation such as Panoptic FPN~\cite{kirillov2019panoptic} and K-net~\cite{zhang2021k} are usually sub-optimal on instance and semantic segmentation compared with specialized models.

Recently, Vision Transformers~\cite{vaswani2017attention} has achieved tremendous success in many computer vision tasks, such as image classification~\cite{liu2021swin,dosovitskiy2020image} and object detection~\cite{carion2020end}. Inspired by DETR's set prediction mechanism, Mask2Former\cite{cheng2021mask2former} proposes a unified framework and achieves a remarkable performance on all three segmentation tasks. Similar to DETR, it uses learnable queries in Transformer decoders to probe features and adopts bipartite matching to assign predictions to ground truth (GT) masks. Mask2Former also uses predicted masks in a decoder layer as attention masks for the next decoder layer. The attention masks serve as a guidance to help next layer's prediction and greatly ease training.

Despite its great success, Mask2Former is still an initial attempt which is mainly based on a vanilla DETR model that has a sub-optimal performance compared to its later variants. For example, in each decoder layer, the predicted masks are built from scratch by dot-producting queries and feature map without performing layer-wise refinement as proposed in deformable DETR\cite{zhu2020deformable}. Because each layer build masks from scratch, the masks predicted by a query in different layers may change dramatically. To show this inconsistent predictions among layers, we build two metrics $\text{mIoU-L}^i$ and $\text{Util}^i$ to quantify this problem and analyze its consequences in Section~\ref{failure}. As a result, this problem leads to a low utilization rate of decoder queries, which is especially severe in the first few decoder layers. As the attention masks predicted from an early layer are usually inaccurate, when serving as a guidance for its next layer, they will lead to sub-optimal predictions in the next layer.

Since detection is a similar task to segmentation, both aiming to locate instances in images, we seek inspirations from detection to improve segmentation. For example, the recently proposed denoising training~\cite{li2022dn} for detection has shown very effective to stabilize bipartite matching between training epochs. DINO~\cite{zhang2022dino} achieves a new SOTA on COCO detection built upon an improved denoising training method. The key idea in denoising training is to feed noised GT boxes in parallel with learnable queries into Transformer decoders and train the model to denoise and recover the GT boxes. 

Inspired by DN-DETR~\cite{li2022dn}, we propose a mask-piloted (MP) training approach. We divide the decoder queries into two parts, an MP part and a matching part. The matching part is the same as in Mask2Former, feeding learnable queries and adopting bipartite matching to assign predictions to GT instances. In the MP part, we feed class embeddings of GT categories as queries and GT masks as attention masks into a Transformer decoder layer and let the model to reconstruct GT masks and labels. To reinforce the effectiveness of our method, we propose to apply MP training for all decoder layers. Moreover, we also add point noises to GT masks and flipping noises to GT class embeddings to force the model to recover GT masks and labels from inaccurate ones.
\par
We summarize our contributions as follows.
\begin{enumerate}
    \item We propose a mask-piloted training approach to improving masked-attention in Mask2Former, which suffers from inconsistent and inaccurate mask predictions among layers. We also develop techniques including multi-layer mask-piloted training with point noises and label-guided training to further improve the segmentation performance. Our method is computationally efficient, introducing no extra computation during inference and little computation during training.
    \item 
    Through analyzing the predictions of Mask2Former, we discover a failure mode of Mask2Former---inconsistent predictions between consecutive layers. We build two novel metrics to measure this failure, which are layer-wise query utilization and mean IoU. We also show that our method improves the two metrics by a large margin, which validates its effectiveness on alleviating inconsistent predictions.
    \item When evaluating on three datasets, including ADE20k, Cityscapes, and MS COCO, our \M outperforms Mask2Former by a large margin. Besides improved performance, our method also speeds up training. Our model exceeds Mask2Former on all three segmentation tasks with only half of the number of training steps on ADE20k. Also, our model trained for $36$ epochs is comparable with Mask2Former trained for $50$ epochs on instance and panoptic segmentation on MS COCO. 
    
\end{enumerate}
\section{Related Work}
Generally, image segmentation can be divided into three tasks including instance segmentation, semantic segmentation, and panoptic segmentation with respect to different semantics.\\\noindent\textbf{Traditional Segmentation models}
 Traditionally, researchers develop specialized models and optimization objectives for each task. Instance segmentation is to predict a set of binary masks and their associated categories. Previous methods often predict masks based on bounding boxes produced by detection models. Mask R-CNN~\cite{he2017mask} builds segmentation upon detection model Faster R-CNN~\cite{ren2015faster} by adding a mask branch in parallel with the detection branch. HTC~\cite{chen2019hybrid} further proposes to interleave the two branches and add mask information flow to improve segmentation performance. Semantic segmentation focus on category-level semantics without distinguishing instances. Previous models widely formulate it into a per-pixel classification problem. The pioneering work, FCN~\cite{long2015fully} generates a label for each pixel to solve this problem. Many follow-up works continue with this idea and design more precise pixel-level classification models~\cite{chen2017deeplab,chen2017rethinking}. Panoptic segmentation~\cite{kirillov2019panoptic,wang2021max} is a combination of the two segmentation tasks above to segment both the foreground instances ("thing") and background semantics ("stuff"). 
\\\noindent\textbf{Vision Transfomers for segmentation}
Transformer-based segmentation models emerge with DETR (DEtection TRansformer)~\cite{carion2020end}. Since that, Transformer~\cite{vaswani2017attention} has been widely used in many detection~\cite{carion2020end,liu2022dab, li2022dn,zhang2022dino} and segmentation models~\cite{cheng2021maskformer, cheng2021mask2former, xie2021segformer}. Generally, these models adopt a set prediction objective with bipartite matching to query features of interest. Inspiring progress has been made in many specialized query-based architectures for instance segmentation~\cite{fang2021instances}, semantic segmentation~\cite{xie2021segformer, zheng2021rethinking}, panoptic segmentation~\cite{li2021panoptic}. Notably, some recent works unify the three segmentation tasks into one model~\cite{zhang2021k, li2022mask, cheng2021maskformer,cheng2021mask2former}. Among them, Mask2Former~\cite{cheng2021mask2former} proposes masked attention and achieves remarkable performance on all three segmentation tasks. Following DETR, Mask2Former is an end-to-end Transformer for image segmentation. It is composed of a backbone, a pixel decoder, and a Transformer decoder. The pixel decoder is an encoder-only deformable Transformer that takes multi-scale image features as input and output processed multi-scale features. The Transformer decoder takes learnable queries as input to attend to the feature maps output by the pixel decoder. Each decoder layer attends to a feature map of one scale following a coarse-to-fine manner. The predicted masks of one decoder layer will feed to the next layer as the attention masks, which enables the following decoder to focus on more meaningful regions. The design of attention masks eases the training and improves performance.
\\\noindent\textbf{Ease Training for Vision transformers}
Vision transformers have an advantage over the convolution-based methods due to their global attention--- they can see all parts of the image. However, it is hard to train because globally searching for an object is a hard task. This phenomenon exists in both detection and segmentation. In detection, DETR suffers from slow convergence requiring $500$ training epochs for convergence. Recently, researchers have dived into the meaning of the learnable queries~\cite{meng2021conditional,zhu2020deformable,wang2021anchor,liu2022dab}. They either express the queries as reference points or anchor boxes. \cite{li2022dn,zhang2022dino} proposed to add noised ground truth boxes as positional queries for denoising training and they speed up detection greatly. In segmentation, Mask2Former proposed mask attention which makes training easier and speeds up convergence when compared with MaskFormer. We have found some common points for these works that speed up training for vision Transformers. Firstly, they usually give clearer meaning to learnable queries to reduce ambiguity. Secondly, they give local limitations to cross attention and reduce the scope for the model to search objects. Therefore, we believe that giving guidance to cross attention leads to good segmentation performance. Our method is similar to DN-DETR~\cite{li2022dn} in introducing GT annotations to the Transformer decoder but ours is totally different from it in 
both the problems we solve and the method we take. Firstly, our method aims to address the inconsistent prediction problem which leads to the low utilization of queries, while DN-DETR aims to solve unstable matching among epochs. Secondly, we use GT masks as the attention masks, while DN-DETR feeds noised GT boxes as decoder queries. Thirdly, noises are optional in our method, but DN-DETR does not work without noises. 

\section{Failure Cases and Solution}
\label{failure}
 \begin{figure*}[ht]
\centering
\subfloat[]{
\begin{minipage}[h]{0.2\linewidth}
\centering
\includegraphics[width=0.75in]{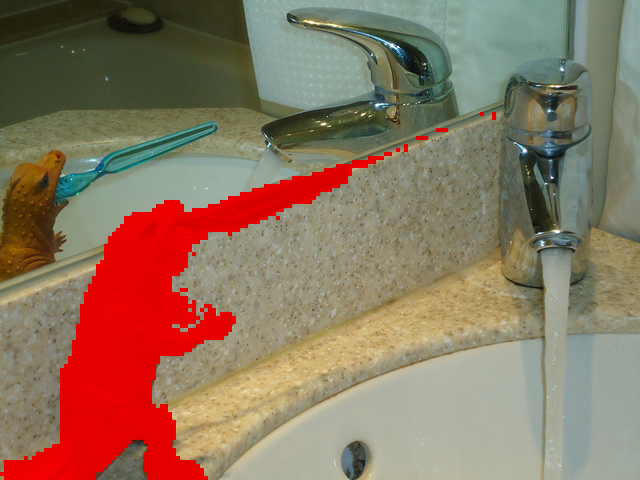}
\end{minipage}%
}%
\hspace{-0.15in}
\subfloat[]{
\begin{minipage}[h]{0.2\linewidth}
\centering
\includegraphics[width=0.75in]{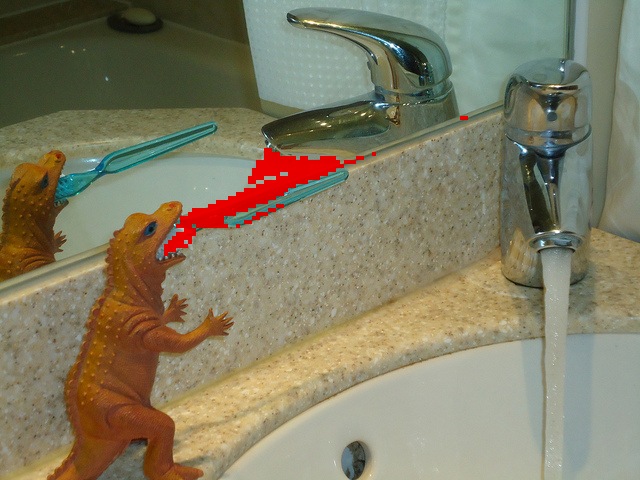}
\end{minipage}%
}%
\hspace{0.25in}
\subfloat[]{
\begin{minipage}[h]{0.2\linewidth}
\centering
\includegraphics[width=0.75in]{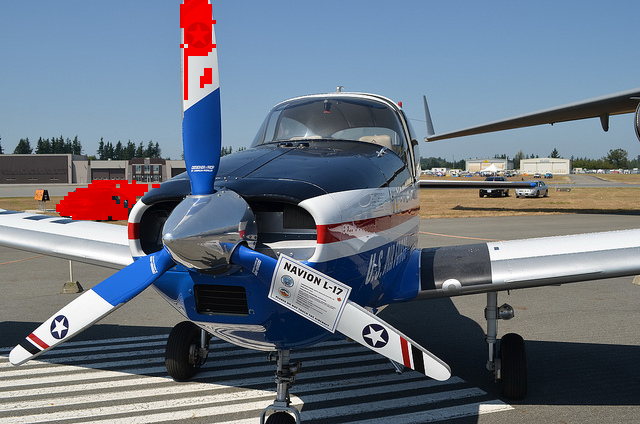}
\end{minipage}
}
\hspace{-0.15in}
\subfloat[]{
\begin{minipage}[h]{0.2\linewidth}
\centering
\includegraphics[width=0.75in]{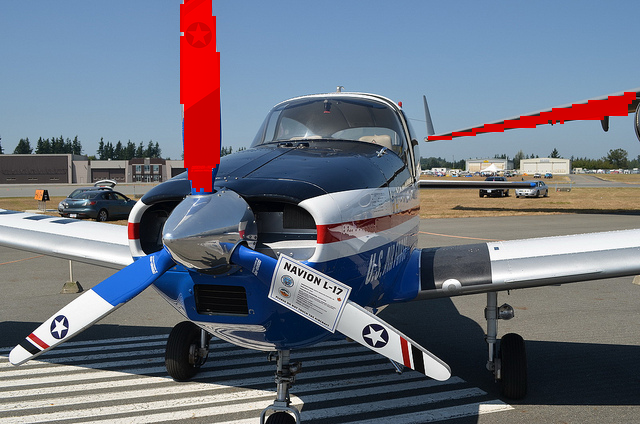}
\end{minipage}%
}%

\subfloat[]{
\begin{minipage}[h]{0.20\linewidth}
\centering
\includegraphics[width=0.75in]{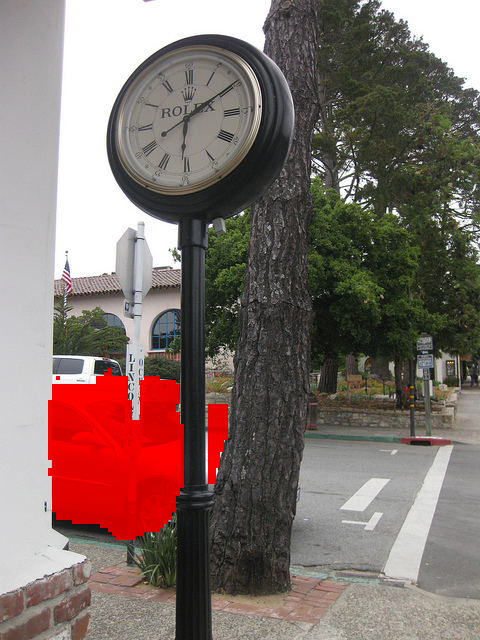}
\end{minipage}%
}%
\hspace{-0.15in}
\subfloat[]{
\begin{minipage}[h]{0.20\linewidth}
\centering
\includegraphics[width=0.75in]{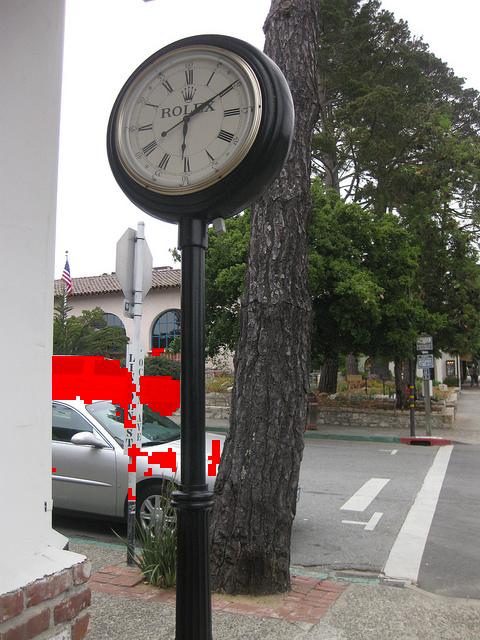}
\end{minipage}
}
\hspace{0.15in}
\subfloat[]{
\begin{minipage}[h]{0.20\linewidth}
\centering
\includegraphics[width=0.75in]{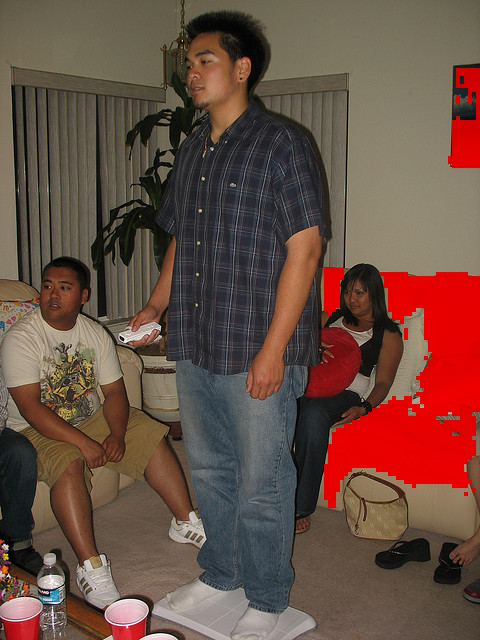}
\end{minipage}%
}%
\hspace{-0.15in}
\subfloat[]{
\begin{minipage}[h]{0.20\linewidth}
\centering
\includegraphics[width=0.75in]{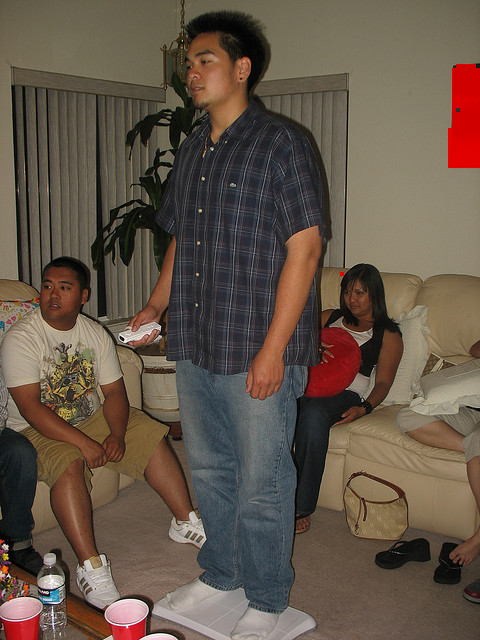}
\end{minipage}%
}%

\centering
\caption{We visualize layer-wise predictions of Mask2Former and show $4$ pairs of failure cases. Each pair is the predictions of the same query in adjacent decoder layers. The red regions are the predicted masks. These cases shows that the predictions of a query may change dramatically between consecutive layers.}
\label{fig:failuer}
\end{figure*}

 \begin{table*}[]
\tablestyle{4pt}{1.2}\scriptsize\begin{tabular}[ht]{lllllllllll}
\midrule
 Layer Number                            &   & 1 & 2 & 3 & 4 & 5 & 6 & 7 & 8&9 \\\midrule
\multirow{2}{*}{Mask2Former}& $\text{mIoU-L}^i$(\%) & 23.0 & 50.3 & 59.5 & 69.7 & 70.1 & 69.9 & 78.6 & 76.3 & 76.0 \\ & $\text{Util}^i$(\%)  & 38.1 & 65.5 & 70.8 & 73.5 & 78.5 & 81.1 & 83.2 & 83.2 &100  \\
                             \midrule
\multirow{2}{*}{MP training} & $\text{mIoU-L}^i$(\%)  & 51.5  & 83.2 & 86.6 & 96.2 & 96.0 & 93.8 & 97.6 & 97.9 & 96.3\\       & $\text{Util}^{i*}$(\%)  & 94.0 & 93.8 & 94.6 & 93.2 & 94.0 & 94.8 & 94.0 & 94.0&100 \\
                             \midrule
\multirow{2}{*}{Multi-layer MP}  & $\text{mIoU-L}^i$(\%)  & 51.3 & 87.8 & 88.8 & 97.5 & 97.5 & 96.2 & 99.0 & 98.8 & 98.2\\      & $\text{Util}^{i*}$(\%)  & 94.2 & 95.4 & 97.3 & 97.6 & 97.3 & 97.0 & 98.2 & 98.2&100 \\
                             \midrule
\end{tabular}
\caption{The $\text{mIoU-L}^i$ and $\text{Util}^i$ of Mask2Former and the proposed methods. The $\text{mIoU-L}^i$ for a query is taken as the IoU of the predicted masks in the $(i-1)$-th and $i$-th decoder layers. The $\text{mIoU-L}^i$ takes the average over all queries. Note that the index of decoder layers starts from $1$, so "prediction of the $0$-th decoder" means the prediction before entering Transformer decoder. The $\text{Util}^i$ is the ratio of GT masks that match with the same query in $i$-th layer and the last layer. $^*$ 
means that We test $\text{Util}^i$ with bipartite matching to assign predictions in MP part to GT masks. Note that when directly assigning predictions in MP part to corresponding GT masks with hard assignment, $\text{Util}^i$ is always $100\%$.}
\label{tab:util}
\vspace{-0.5cm}
\end{table*}
\noindent\textbf{Failure cases:} We view the mask prediction in Mask2Former as a mask refinement process through Transformer decoder layers. By taking predictions from a layer as the attention masks of the next layer, we expected to obtain refined predictions from the next layer. Generally, the predicted masks are assigned to the GT masks by bipartite matching and refined in a coarse-to-fine procedure. Therefore, keeping refinement consistency in different layers is the key to refining the predicted masks consistently. However,
when we visualize the predictions of Mask2Former layer by layer, we found a severe inconsistency problem among the predictions from the same query between consecutive decoder layers. As shown in Fig.~\ref{fig:failuer}, the predictions of adjacent layers differ significantly. We define a metric named layer-wise mean Intersection-over-Union (IoU) denoted as $\text{mIoU-L}^i$ to quantify this inconsistency. We denote the predicted masks from the $i$-th decoder layer as $\mathbf{M^i}=\left\{M_0^i, M_1^i, ..., M_{N-1}^i\right\}$, where $N$ is the number of decoder queries. Note that $M^0$ means the predictions before queries enter the first Transformer decoder layer. We have
 \begin{equation}
     \text{mIoU-L}^i=\sum_{n=0}^{N-1}\text{IoU}(M_n^{i-1},M_n^{i})
 \end{equation}
As shown in Table~\ref{tab:util}, the mIoU-L is quite low, especially for the first few layers, which indicates inconsistent mask prediction results between adjacent layers. Such an inconsistent result means a query may be assigned to different instances (or background for panoptic and semantic segmentation) during layer-by-layer updates. To evaluate the inconsistent matching problem, we define another metric named as layer-wise query utilization denoted as $\text{Util}^i$ for the $i$-th layer, which means the ratio of the matched queries in the $i$-th layer that matches exactly the same instance as in the last layer. We denote the ground truth objects as $\mathbf{T}=\left\{T_0, T_1, T_2, ..., T_{O-1}\right\}$ where $O$ is the number of ground truth objects. After bipartite matching, we compute an index vector $\mathbf{V^i}=\left\{V^i_0, V_1^i, ..., V_{N-1}^i\right\}$ to store the matching result of layer $i$ as follows.
\begin{equation}
    V^i_n=\left\{\begin{array}{ll}
 o, &\text{if } M^i_n \text{ matches } T_o\\
 -1, &\text{if } M^i_n \text{ matches nothing}
\end{array}\right.
\label{eq: matching}
\end{equation}
We define the query utilization of layer $i$ as the proportion of GT masks that match with the same query in layer $i$ and the last layer, which is calculated as
\begin{equation}
    \text{Util}^i=\frac{1}{O}\sum_{j=0}^{N}\mathbbm{1}(V^i_n= V^{l}_n \ \text{and} \  V^i_n\neq -1),
    \label{eq: instability}
\end{equation}
where $\mathbbm{1}(\cdot)$ is the indicator function. $\mathbbm{1}(x)=1$ if $x$ is true and $0$ otherwise. The query utilization of layer $i$ for the whole data set is averaged over the utilization numbers for all images. Table~\ref{tab:util} shows that the query utilizations of the first few layers are too low, which indicate many matched queries in early layers are ended up not being used in the last layer. The low $\text{mIoU-L}^i$ and $\text{Util}^i$ have 
potential consequences such as inconstant optimization goals and low training efficiency.

Our observations based on these two metrics motivate us to develop a more effective training method to make the matching more consistent among layers.

\noindent\textbf{Solution:} To address the inconsistent prediction problem, inspired by DN-DETR, we proposed a mask-piloted training method. We feed GT masks as attention masks and expect these queries with GT masks can focus on their corresponding GT masks and not be distracted by other instances. As shown in row 2 and 3, Table~\ref{tab:util}, both $\text{mIoU-L}^i$ and $\text{Util}^i$ are improved by a large margin when adopting mask-piloted training.
We also find that only applying mask-piloted (MP) training in the first layer is not enough. Fig~\ref{fig:m2q}(a)(b) shows that inconsistent prediction exists even when given GT masks. Therefore, we add a GT mask to all decoder layers to reinforce MP training. As shown in Table~\ref{tab:util}, multi-layer MP training further improves both $\text{mIoU-L}^i$ and $\text{Util}^i$ especially in latter decoder layers.




\section{\M}

\subsection{Overview}
\begin{figure*}[h]
\vspace{-0.3cm}
\centering
\begin{minipage}[h]{1\linewidth}
\centering
\includegraphics[width=0.8\linewidth]{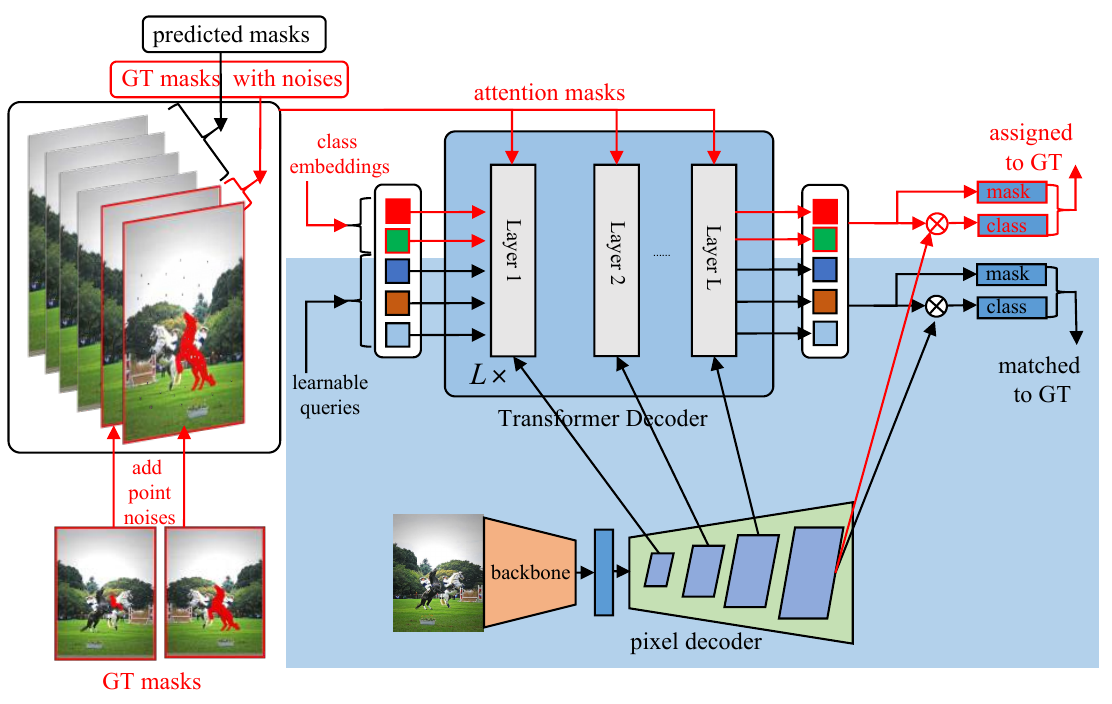}
\end{minipage}%
\caption{The architecture of our method is the same as Mask2Former (the blue-shaded part), which consists of a backbone, a pixel decoder, and a Transformer decoder. The difference is that we feed extra queries and attention masks which are called the MP part to the Transformer decoder (red-line part in the figure). The MP part contains GT masks as attention masks and GT class embeddings as queries. We feed GT masks into the MP part of all decoder layers. We also add point noises to GT masks and flipping noises to class embeddings which can further improve the performance. Note that this architecture is just for training. In the inference time, the red-line part does not exist, and thus, our pipeline is exactly the same as Mask2Former.}
\label{fig:arch}
\end{figure*}

Our method is a training method for improving Mask2Former.
Our \M only introduces minor changes in the Mask2Former pipeline during training as shown in the red-line part in Fig.~\ref{fig:arch}. Same as Mask2Former, our model is composed of a backbone, a pixel decoder, and a Transformer decoder. Our improvements are on the Transformer decoder. We divide decoder queries into two parts, a mask-piloted (MP) part and a matching part. The matching part is the same as Mask2Former. The MP part feeds GT masks and GT class embeddings as input and the predictions are directly assigned to the corresponding GT masks without bipartite matching. Our method includes three components which are multi-layer MP training, mask noises, and label-guided training.
\subsection{Multi-layer MP training}
 In order to address the inconsistent prediction problem, we propose a multi-layer mask-piloted (MP) training method. In the MP part, we feed an additional set of queries and masks into Transformer where the queries and masks are GT class embeddings and GT masks. On the loss side, we independently assign the outputs of the MP part and the matching part with the GT instances. The predictions of the MP part are directly assigned to their corresponding GT masks and those in the matching part are assigned with bipartite graph matching. The losses for both parts are the same, following the loss design in Mask2Former. We also propose to add GT masks in multiple layers and show that this can further improve the performance with experiments in Table~\ref{tab:num layers}. Since we follow Mask2Fomer to use feature maps of different resolutions in different decoder layers, we interpolate GT masks into different resolutions when applied to different layers. 
\begin{figure*}[h]

    \centering

\centering

\subfloat[]{
\begin{minipage}[h]{0.16\linewidth}
\centering
\includegraphics[width=1.\linewidth]{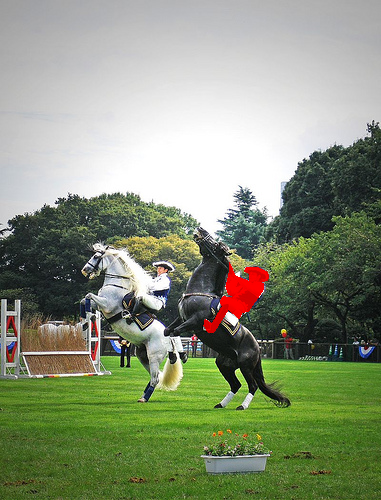}
\end{minipage}%
}%
\centering
\subfloat[]{
\begin{minipage}[h]{0.16\linewidth}
\centering
\includegraphics[width=1.\linewidth]{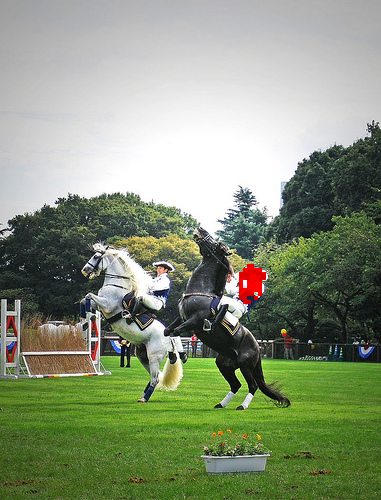}
\end{minipage}%
}%
\centering
\subfloat[]{
\begin{minipage}[h]{0.16\linewidth}
\centering
\includegraphics[width=\linewidth]{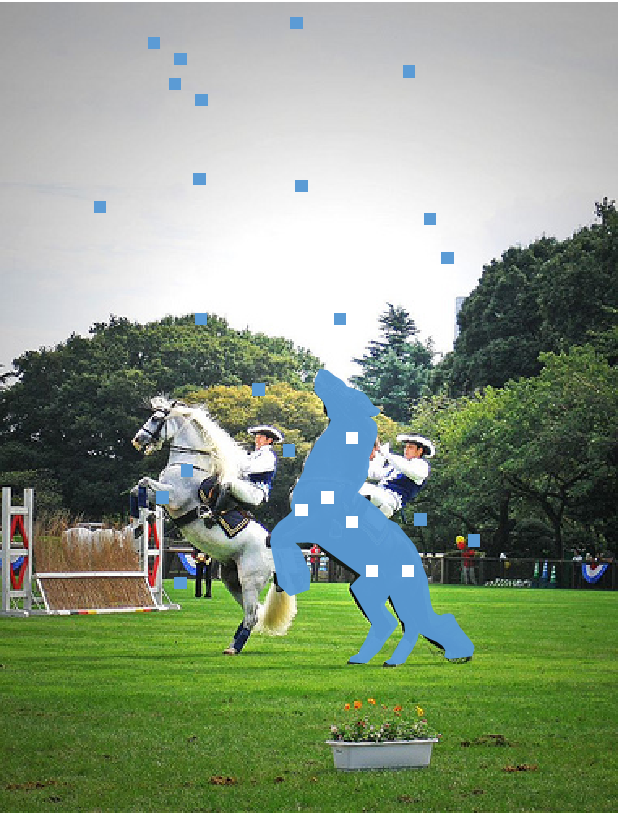}
\end{minipage}%
}%
\centering
\subfloat[]{
\begin{minipage}[h]{0.16\linewidth}
\centering
\includegraphics[width=\linewidth]{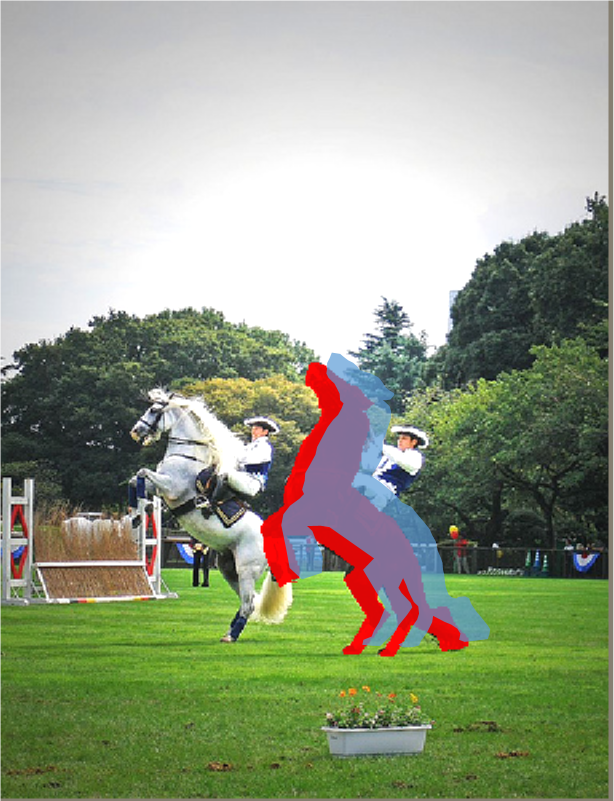}
\end{minipage}%
}%
\centering
\subfloat[]{
\begin{minipage}[h]{0.16\linewidth}
\centering
\includegraphics[width=\linewidth]{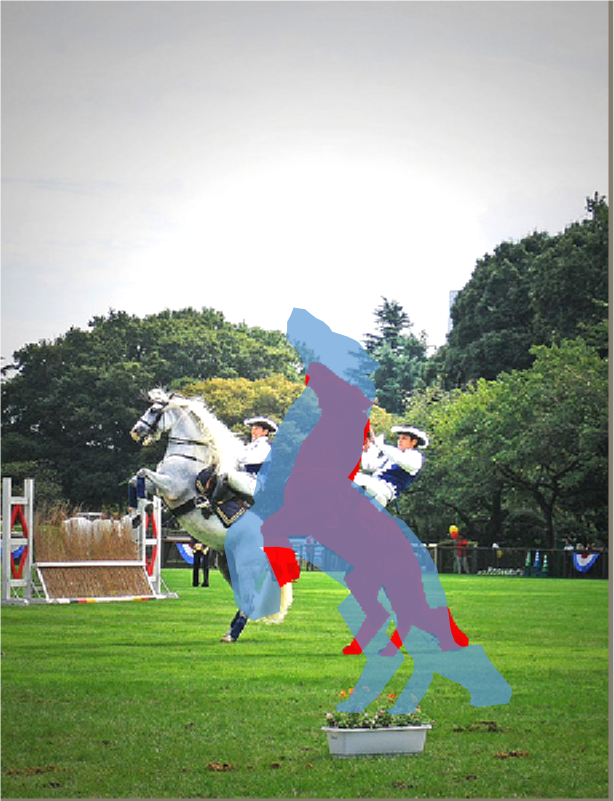}
\end{minipage}%
}%
\caption{(a)The red region is a given GT mask in first decoder layer which covers the man riding the black horse. (b) The predicted mask of the first decoder layer which only covers the head of the man. The predicted mask will be used as the attention mask of the second layer and may mislead the second layer. (c)(d)(e) are demonstrations of point, shift and scale noises. The blue regions are noised masks and the red regions are the GT masks.}
\end{figure*}
\begin{table*}[!h]
  \centering

  \tablestyle{3pt}{1.2}\scriptsize\begin{tabular}{l|l | x{28}x{28}x{28} | x{20}x{20}x{20}x{20} |x{36}x{36}}
  \multirow{2}{*}{method} & \multirow{2}{*}{backbone} & \multicolumn{3}{c|}{panoptic } & \multicolumn{4}{c|}{instance } & semantic  \\
   &  & PQ & AP$^\text{Th}_\text{pan}$ & mIoU$_\text{pan}$ & AP & AP$^\text{S}$ & AP$^\text{M}$ & AP$^\text{L}$ & mIoU \\
  \shline
  MaskFormer~\cite{cheng2021maskformer} & R50 & 34.7\phantom{$^*$} & - & - & - & - & - & - & -  \\
  Panoptic-DeepLab~\cite{cheng2020panoptic} & SWideRNet~\cite{chen2020scaling} & 37.9$^*$ & - & 50.0$^*$ & - & - & - & - & -  \\
  Mask2Former~\cite{cheng2021mask2former} & R50\phantom{$^{\text{\textdagger}}$}
  & 39.7\phantom{$^*$} & 26.5 & 46.1\phantom{$^*$} & 26.4 & 10.4 & 28.9 & 43.1 & 47.2 \\
  \textbf{\M} & R50\phantom{$^{\text{\textdagger}}$}
  & \textbf{40.8(+1.1)} & \textbf{27.1} & \textbf{48.3(+2.2)} & \textbf{28.0(+1.6)} & 10.5 & \textbf{30.7(+1.8)} & \textbf{44.6} & \textbf{48.1} \\
  \hline
  MaskFormer~\cite{cheng2021maskformer} & Swin-L$^{\text{\textdagger}}$ & - & - & - & - & - & - & - & 54.1 \\
  FaPN-MaskFormer~\cite{huang2021fapn,cheng2021maskformer} & Swin-L$^{\text{\textdagger}}$ & - & - & - & - & - & - & - & 55.2  \\
  Mask2Former~\cite{cheng2021mask2former}& Swin-L$^{\text{\textdagger}}$ & 48.1\phantom{$^*$} & 34.2 & 54.5\phantom{$^*$} & 34.9 & 16.3 & 40.0 & 54.7 & 56.1 \\
  \textbf{\M} & Swin-L$^{\text{\textdagger}}$ & \textbf{49.4(+1.3)} & \textbf{35.2} &\textbf{56.0}  & \textbf{35.3} & 16.3 & \textbf{40.7} & \textbf{55.5} & \textbf{56.9}  \\

  \end{tabular}

   \caption{\textbf{Image segmentation results on ADE20K \texttt{val}.}  \M improves the performance on all three segmentation tasks with R$50$ and Swin-L as backbone. All metrics are evaluated with \emph{single-scale} inference. Backbones pre-trained on ImageNet-22K are marked with $^{\text{\textdagger}}$. - means the results are not reported for the methods.}

\label{tab:benchmark:ade20k}
\vspace{-0.2in}
\end{table*}
\begin{table*}[h]
  \centering

  \tablestyle{4pt}{1.2}\scriptsize\begin{tabular}{l|l | x{28}x{28}x{28} | x{28}x{28} |x{36}x{36}}
  \multirow{2}{*}{method} & \multirow{2}{*}{backbone} & \multicolumn{3}{c|}{panoptic } & \multicolumn{2}{c|}{instance } & {semantic } \\
   &  & PQ & AP$^\text{Th}_\text{pan}$ & mIoU$_\text{pan}$ & AP & AP50 & mIoU\\
  \shline
Panoptic-DeepLab~\cite{cheng2020panoptic}
  & R50 & 60.3  & 32.1 & 78.7 & - & - & -  \\
  Mask2Former~\cite{cheng2021mask2former}
  & R50& 62.1  & 37.3 & 77.5 & 37.4 & 61.9 & 79.4  \\
  \textbf{\M} & R50 & \textbf{62.7}  & \textbf{39.3(+2.0)} & \textbf{79.6(+2.1)} & \textbf{39.7(+2.3)} & \textbf{65.4(+3.5)} & \textbf{81.0(+1.6)}  
  \\\hline
 Mask2Former~\cite{cheng2021mask2former} & Swin-L$^{\text{\textdagger}}$ & 66.6  & 43.6 & 82.9 & 43.7 & 71.4 & 83.3
 \\
 \textbf{\M} & Swin-L$^{\text{\textdagger}}$ & \textbf{67.5}  & \textbf{45.8(+2.2)} & \textbf{83.5} & \textbf{44.9(+1.2)} & \textbf{72.4} & \textbf{83.9}  \\
  \end{tabular}

  \caption{\textbf{Image segmentation results on Cityscapes \texttt{val}.} We report results on all three segmentation tasks with R50 or Swin-L as backbone. All metrics are evaluated with \emph{single-scale} inference. Backbones pre-trained on ImageNet-22K are marked with $^{\text{\textdagger}}$}

\label{tab:benchmark:cityscapes_full}
\end{table*}
\subsection{Noised masks}
Besides feeding GT masks to alleviate the inconsistency problem, we further propose to use noised GT masks to make mask refinement more robust in each decoder. The intuition is simple, as Mask2former refines the mask from an inaccurate mask predicted from the previous decoder. GT masks without noise may be too easy for the task which prevents further refinement. Therefore, we feed noised masks to the decoder and train the model to reconstruct the original ones, which is similar to DN-DETR~\cite{li2022dn} for box denoising training.
We have tried three types of noise which are point noise, shift noise, and scale noise as shown in Fig.~\ref{fig:m2q}. For point noise, we randomly sample some points on the mask and flip them from 1 to 0 or from 0 to 1. The number of sampled points is proportional to the area of a GT mask. For shift noise, we randomly shift a mask instance vertically and horizontally while keeping the center of the instance within its GT bounding box. For the scaling noise, we enlarge or shrink the instance by a random ratio. A comparison of three types of noises is shown in Table~\ref{tab:num layers}. Through experiments, we find point noises improve the performance while shifting and scaling noises do not work for our method. Therefore, we adopt point noises in our implementation. We include a discussion on noise types in Appendix~\ref{supp:noise type}. We use a noise ratio $\lambda_p$ to control how many noise points are added. Denote a binary GT mask as $M=\left[m_{i,j}\right]_{H\times W}$ where $m_{i,j}=1$ means the pixel belongs to the mask. The number of noise points is randomly sampled from $0$ to $\lambda_p \times\sum_{i,j}m_{i,j}$. Since we apply mask-piloted training to all decoder layers, we independently add different noises to GT masks in different layers.
\begin{table*}[h]
  \centering

  \tablestyle{4pt}{1.2}\scriptsize\begin{tabular}{l| lcc | x{20}x{20}x{20}x{20} |x{24}x{24}x{18}}
  method & backbone & query type & epochs & AP & AP$^\text{S}$ & AP$^\text{M}$ & AP$^\text{L}$  & \#params. & FLOPs & fps \\
  \shline
  MaskFormer~\cite{cheng2021maskformer} & R50 & 100 queries & 300 & 34.0 & 16.4 & 37.8 & 54.2  & \phantom{0}45M & \phantom{0}181G & 19.2 \\
   {Mask R-CNN~\cite{he2017mask}} &  {R50} &  {dense anchors} &  {36} &  {37.2} &  {18.6} &  {39.5} &  {53.3}  &  {\phantom{0}44M} &  {\phantom{0}201G} &  {15.2} \\
  Mask R-CNN~\cite{he2017mask,ghiasi2021simple,du2021simple} & R50 & dense anchors & 400 & 42.5 & 23.8 & 45.0 & 60.0  & \phantom{0}46M & \phantom{0}358G & 10.3 \\
  Mask2Former~\cite{cheng2021mask2former}  & R50 & 100 queries & 50 &43.7 & 23.4 & 47.2& 64.8  & \phantom{0}44M & \phantom{0}226G & \phantom{0}9.7 \\
  \textbf{\M}  & R50 & 100 queries & 50 &\textbf{44.8(+1.1)} & \textbf{24.7} & \textbf{48.1}& \textbf{65.5}  & \phantom{0}44M & \phantom{0}226G & \phantom{0}9.7 \\
  \hline
   {Mask R-CNN~\cite{he2017mask}} &  {R101} &  {dense anchors} &  {36} &  {38.6} &  {19.5} &  {41.3} &  {55.3}  &  {\phantom{0}63M} &  {\phantom{0}266G} &  {10.8} \\
  Mask R-CNN~\cite{he2017mask,ghiasi2021simple,du2021simple} & R101 & dense anchors & 400 & 43.7 & \textbf{24.6} & 46.4 & 61.8 & \phantom{0}65M & \phantom{0}423G & \phantom{0}8.6 \\
  Mask2Former~\cite{cheng2021mask2former}  & R101 & 100 queries & 50 & 44.2 & 23.8 &47.7 & \textbf{66.7}  & \phantom{0}63M & \phantom{0}293G & \phantom{0}7.8 \\
  \textbf{\M}  & R101 & 100 queries & 50 & \textbf{45.1(+0.9)} & \textbf{24.6} & \textbf{48.7} & 66.2  & \phantom{0}63M & \phantom{0}293G & \phantom{0}7.8 \\
  \hline
  QueryInst~\cite{fang2021instances}&Swin-L$^\dag$&300 queries &50&48.9& 30.8& 52.6& 68.3 &-&-&3.3 \\
  Swin-HTC++~\cite{liu2021swin,chen2019hybrid}&Swin-L$^\dag$&dense anchors&72&49.5 &31.0& 52.4& 67.2 &284M&1470G&- \\
  Mask2Former~\cite{cheng2021mask2former}&Swin-L$^\dag$&200 queries&100&50.1& 29.9& 53.9 &72.1&216M&868G&4.0 \\
  \textbf{\M}(ours)&Swin-L$^\dag$&200 queries&72&\textbf{50.8(+0.7)}& 30.3& \textbf{54.5} &\textbf{72.6} &216M&868G&4.0 \\
  \end{tabular}
  \vspace{-1mm}

   \caption{\textbf{Instance segmentation on COCO \texttt{val2017} with 80 categories.} \M outperforms strong baseline Mask2Former without extra computation cost in inference time. For a fair comparison, we only consider single-scale inference and models trained using only COCO \texttt{train2017} set data. Backbones pre-trained on ImageNet-22K are marked with $^{\text{\textdagger}}$}

\vspace{-3mm}

\label{tab:insseg:coco}
\end{table*}
\begin{table*}[h]
  \centering

  \tablestyle{3pt}{1.2}\scriptsize\begin{tabular}{l | lcc | x{20}x{20}x{20} | x{20}x{24} |x{24}x{24}x{18}}
  method & backbone & query type & epochs & PQ & PQ$^\text{Th}$ & PQ$^\text{St}$ & AP$^\text{Th}_\text{pan}$ & mIoU$_\text{pan}$ & \#params. & FLOPs & fps \\
  \shline
   {DETR~\cite{carion2020end}} &  {R50} &  {100 queries} &  {500+25} &  {43.4} &  {48.2} &  {36.3} &  {31.1} &  {-} &  {-} &  {-} &  {-} \\
  MaskFormer~\cite{cheng2021maskformer} & R50 & 100 queries & 300 & 46.5 & 51.0 & 39.8 & 33.0 & 57.8 & \phantom{0}45M & \phantom{0}181G & 17.6 \\
  Mask2Former~\cite{cheng2021mask2former} & R50 & 100 queries & 50 & $51.4^*$ & 57.2 & 42.7 & 41.9 & 61.5 & \phantom{0}44M & \phantom{0}226G & \phantom{0}8.6 \\
  \textbf{\M} (ours) & R50 & 100 queries & 50 & \textbf{52.2(+0.8)} & \textbf{58.3} & \textbf{42.9} & \textbf{42.5} & \textbf{61.6} & \phantom{0}44M & \phantom{0}226G & \phantom{0}8.6\\
  \hline
   {DETR~\cite{carion2020end}} &  {R101} &  {100 queries} &  {500+25} &  {45.1} &  {50.5} &  {37.0} &  {33.0} &  {-} &  {-} &  {-} &  {-} \\
  MaskFormer~\cite{cheng2021maskformer} & R101 & 100 queries & 300 & 47.6 & 52.5 & 40.3 & 34.1 & 59.3 & \phantom{0}64M & \phantom{0}248G & 14.0 \\
  Mask2Former~\cite{cheng2021mask2former} & R101 & 100 queries & 50 & 52.6 & 58.5 & 43.7 & 42.6 & \textbf{62.4} & \phantom{0}63M & \phantom{0}293G & \phantom{0}7.2 \\
  \textbf{\M} (ours) & R101 & 100 queries & 50 & \textbf{52.9(+0.3)} & \textbf{58.7} & \textbf{44.0} & \textbf{43.0} & 61.8 & \phantom{0}63M & \phantom{0}293G & \phantom{0}7.2 \\
  \hline
  Max-DeepLab~\cite{wang2021max} & Max-L & 128 queries & 216 & 51.1 & 57.0 & 42.2 & - & - & 451M & 3692G & - \\
  MaskFormer~\cite{cheng2021maskformer} & Swin-L$^{\text{\textdagger}}$ & 100 queries & 300 & 52.7 & 58.5 & 44.0 & 40.1 & 64.8 & 212M & \phantom{0}792G & \phantom{0}5.2 \\
  K-Net~\cite{zhang2021k} & Swin-L$^{\text{\textdagger}}$ & 100 queries & 36 & 54.6 & 60.2 & 46.0 & - & - & - & - & - \\
  Mask2Former & Swin-L$^{\text{\textdagger}}$ & \emph{200 queries} & \emph{100} & 57.8 & 64.2 & 48.1 & 48.6 & 67.4 & 216M & \phantom{0}868G & \phantom{0}4.0 \\
  \textbf{\M} & Swin-L$^{\text{\textdagger}}$ & \emph{200 queries} & \emph{72} & \textbf{58.1(+0.3)} & \textbf{64.4} & \textbf{48.4} & \textbf{48.9} & \textbf{67.6} & 216M & \phantom{0}868G & \phantom{0}4.0 \\
    
  \end{tabular}
  \vspace{-1mm}

   \caption{\textbf{Panoptic segmentation on COCO panoptic \texttt{val2017} with 133 categories.} \M outperforms Mask2Former on COCO panoptic segmentation with both R$50$ and R$101$ as backbone. Results that are implemented are marked with $^*$.
}
\vspace{-2mm}

\label{tab:panseg:coco}
\end{table*}
\subsection{Label-guided training}
We use the class embeddings of GT categories as queries in the MP part. The reason why we use class embeddings is that queries will dot-product with image features and an intuitive way to distinguish instances is to use their categories. Note that we only use class embeddings in the first layer and do not replace predicted queries with class embeddings in subsequent layers. We also have a classification loss in the MP part. 
Since the classification task becomes very easy given GT class embeddings, we add flipping noises in class embeddings. We randomly sample GT class embeddings and let them randomly flip to the class embeddings of other categories. For example, ${E_0, E_1,..., E_K}$ denotes the class embeddings of $K$ categories. An image contains $N$ instances which belong to Categories $C^0, C^1, ..., C^N$. So their default class embeddings should be ${E_{C^0}, E_{C^1}, ...,E_{C^N}}$. To add flipping noise, we enumerate all $N$ instances. For $i$-th instance, we generate a random number $r_i$ in $[0,1]$ to decide whether to flip the class embedding $E_{C^i}$ or not. If $r_i<\lambda_l$, we randomly sample another class embedding in ${E_0, E_1,..., E_K}$ to replace $E_{C^i}$. $\lambda_l$ denotes the proportion of the sampled class embeddings.

\section{Experiments}
We conduct extensive experiments to compare our method with Mask2Former and other segmentation methods on all three segmentation tasks on three datasets to show the effectiveness of our method. We also conduct several ablation studies to show the effectiveness of each component in our method.\\
\vspace{-0.2in}
\subsection{Implementation details}
\noindent\textbf{Model:} Since our method is an improved training method based on Mask2Former, we adopt an identical model setting as Mask2Former except for the mask-piloted (MP) part. In the MP part, we adopt $100$ GT masks with point noises. We adopt noise ratios $\lambda_p=0.2$ and $\lambda_l=0.2$ for point noises on masks and flipping noises on class embeddings. We adopt ResNet$50$~\cite{he2015deep} and Swin-L~\cite{liu2021swin} when evaluating our method on ADE20K and Cityscapes. On COCO2017, we use ResNet$50$ and ResNet$101$ as the backbone.

\textbf{Dataset and metrics: }We evaluate \M on three challenging datasets: Cityscapes~\cite{Cordts2016Cityscapes,Cordts2015Cvprw} and ADE20K~\cite{zhou2019semantic} for all three segmentation tasks and COCO 2017~\cite{lin2015microsoft} dataset for panoptic segmentation and instance segmentation. For instance segmentation, we evaluate the mask AP~\cite{lin2015microsoft} on instances that are denoted as "thing" categories in datasets. For semantic segmentation, we evaluate mean Intersection-over-Union (mIOU) over all categories including both foreground and background.  For panoptic segmentation, we evaluate the results with the panoptic quality (PQ) metric~\cite{kirillov2019panoptic}.

More implementation details are shown in Appendix~\ref{supp:imp detail}.
\begin{table*}[h]
  \centering

  \tablestyle{4pt}{1.2}\scriptsize\begin{tabular}{l|l|l|l | x{15}x{15}x{20}x{15} | x{15}x{15}x{15}x{15}x{15} |x{15}x{15}x{15}}
 \multirow{2}{*}{dataset}&\multirow{2}{*}{method} & \multirow{2}{*}{backbone}& \multirow{2}{*}{epochs$^*$}& \multicolumn{4}{c|}{panoptic } & \multicolumn{5}{c|}{instance } & \multicolumn{2}{c}{semantic}  \\
 &  & & & PQ & AP$^\text{Th}_\text{pan}$ & mIoU$_\text{pan}$&clock time/h & AP & AP$^\text{S}$ & AP$^\text{M}$ & AP$^\text{L}$&clock time/h & mIoU&clock time/h \\
  \shline
  \multirow{4}{*}{ADE20k}&
      Mask2Former & R50\phantom{$^{\text{\textdagger}}$} &160k
      & 39.7 & 26.5 & 46.1& 52.8 & 26.4 & 10.4 & 28.9 & 43.1&40.3 & 47.2& 21.3\\
      &\M & R50\phantom{$^{\text{\textdagger}}$} &\textbf{80k}
      & 39.7 &\textbf{27.0}  & \textbf{48.4} &34.2& \textbf{27.4} & 10.2 & \textbf{30.0} & \textbf{44.5}&25.5 & \textbf{47.4}&12.3 \\
  \cline{2-15}
  
      &Mask2Former& Swin-L$^{\text{\textdagger}}$ &160k& 48.1 & 34.2 & 54.5&63.2& 34.9 & 16.3 & 40.0 & 54.7& 51.4& 56.1& 39.8 \\
      &\M & Swin-L$^{\text{\textdagger}}$ &\textbf{80k}& \textbf{48.5} & \textbf{34.7} & \textbf{56.1}&45.7& \textbf{35.2} & 15.8 & 39.6 & \textbf{55.6}&26.7 & \textbf{56.9}& 21.5 \\
      \hline\hline
\multirow{4}{*}{COCO}&
      Mask2Former & R50 &50
      & $51.4^*$ & \textbf{41.9} & 61.5&77.9 & 43.7 & 23.4 & 47.2 & 64.8&62.5 &-&-\\
      &\M & R50 &\textbf{36}& \textbf{51.5} & 41.8 & \textbf{61.6}&59.5 & \textbf{44.1} & \textbf{24.1} & \textbf{47.4} & \textbf{65.3} &50.8& -&-\\
  \cline{2-15}
      &Mask2Former& R101&50&52.6 & 42.6 & 62.4&113.1 & 44.2 & \textbf{23.8} & 47.7 & \textbf{66.7}&86.0 & -&- \\
      &\M & R101 &\textbf{36}& 52.5\phantom{$^*$} & 42.6 & 62.0 &84.4& 44.2 & 23.2 & \textbf{47.9} & 66.6&71.5 &-&-  \\
  \end{tabular}
   \caption{\textbf{Image segmentation results on ADE20K \texttt{val}. and COCO \texttt{val2017}} \M speeds up training by large. On ADE20K, \M trained for $80$K steps can achieve comparable result with Mask2Former trained for $160$K steps with both R$50$ and Swin-L as backbone. On COCO, \M trained for $36$ epochs can achieve comparable result with Mask2Former trained for $50$ epochs with both R$50$ and R$101$ as backbone. $^*$ means that we adopt number of steps instead of epochs on ADE20k following Mask2Former. We report only single-scale inference results. Backbones pre-trained on ImageNet-22K are marked with $^{\text{\textdagger}}$}

\label{tab:speedup}
\vspace{0.2in}
\end{table*}
\begin{table*}[!t]
\begin{adjustbox}{width=1.0\textwidth,center}
    \begin{minipage}[t]{0.55\textwidth}
\makeatletter\def\@captype{table}
\centering
    \begin{tabular}{lcc}
\#Row          & 12 ep (AP mask) &50 ep (AP mask)\\ \hline\toprule
1. Mask2Former\cite{cheng2021mask2former}    & 38.7 &43.7   \\
2. Row1+MP training on 1st layer    & 39.4  &44.2     \\ 
3. Row2+Label noises   & 39.6  &44.2 \\
4. Row3+Point noises  & 39.9  &44.6 \\
5. Ours (Row4+All-layer MP)    & \textbf{40.2}&\textbf{44.8} \\

\end{tabular}
    \caption{The effectiveness of each component of our method when trained for 12 epochs and 50 epochs. The table shows that each component of our model effectively improves the results. Label noise does not work when trained for 50 epochs. Espectially, our method without noises can improve the results by $+0.9$AP. With multi-layer MP training, our method achives $+1.5$AP on COCO instance segmentation with R50 backbone trained for $12$ epochs. "ep" is short for epochs.
    }
    \label{tab:component}
    \end{minipage}\hspace{10mm}

    \begin{minipage}[t]{0.42\textwidth}
\makeatletter\def\@captype{table}
\centering
\vspace{-45pt}

\begin{tabular}{lll}
\textbf{Layers for MP training}   & ep & AP mask \\ \hline\toprule
First layer   & 12 & 39.4    \\ 
First three layers & 12 & 39.6(+0.2)    \\ 
All layers         & 12 & 40.2(+0.8)    \\
\textbf{Noise Type}   &  &   \\ \hline\toprule
No noise  & 12 & 39.6 \\ 
Point noise  & 12 & \textbf{39.9} \\ 
Shifting noise   & 12 & 39.6    \\ 
Scaling noise & 12 & 39.4    \\ 
\end{tabular}

\caption{Top $3$ lines shows the effectiveness of multi-layer Mask-Piloted (MP) training. The bottom lines of the table shows that shifting and scaling noise do not work for our method. "ep" is short for epochs.}
\label{tab:num layers}
\end{minipage}

\hspace{6pt}
\begin{minipage}[t]{0.48\textwidth}
\makeatletter\def\@captype{table}
\centering
\vspace{-45pt}

\begin{tabular}{lll}
\textbf{Feature map}         & ep & AP mask \\ \hline\toprule
Coarse to fine      & 12 & 40.2    \\ 
All large           & 12 & 40.0    \\ 
\textbf{Assignment in Mask-Piloted part} &    &         \\\hline\toprule
Match                 & 12 & 39.4    \\
Hard assignment                  & 12 & 39.6   
\end{tabular}

\caption{Some hyperparemeter tuning:1.coarse-to-fine manner works better for our method than all-large feature maps.  2.Hard assignment lead to better results than matching for MP part queries. "ep" is short for epochs.}
\label{tab:hyper param}
\end{minipage}

    \end{adjustbox}
\end{table*}

\subsection{Main Results}
\vspace{-0.1in}

We trained on three challenging datasets ADE20K, Cityscapes, and MS COCO with different backbones. Our method outperforms Mask2Former on all three segmentation tasks (instance, panoptic, and semantic) under the same setting, which demonstrates its effectiveness and generalization ability to different datasets. 
\\\textbf{Performance on ADE20K. }
On ADE20K, we report our performance trained for $160K$ steps following Mask2Former. 
As shown in Table~\ref{tab:benchmark:ade20k}, our method exceeds Mask2Former and other baselines on all metrics with both R$50$ and Swin-L. Notably, \M achieves $+1.6$AP on instance segmentation with a R$50$ backbone and $+1.3$PQ on panoptic segmentation with a Swin-L backbone. 
\\\textbf{Performance on Cityscapes. }
On Cityscapes, we train the model with $90$K training steps following Mask2Former. 
As shown in Table~\ref{tab:benchmark:cityscapes_full}, our method consistently outperforms Mask2Former on all three segmentation tasks across different backbones. \M also scales well when adopting Swin-L as the backbone. Especially, our method outperforms Mask2Former by $+2.3$AP on instance segmentation and $+1.6$mIoU on semantic segmentation with R$50$. 
\\\textbf{Performance on COCO. }
On COCO val2017, we follow Mask2Former's setting to train for $50$epochs. As shown in Table~\ref{tab:insseg:coco} and Table~\ref{tab:panseg:coco}, \M outperforms both Mask2Former and classical methods such as Mask R-CNN\cite{he2017mask} on instance segmentation and panoptic segmentation. For instance segmentation, our method achieves $+1.1$AP with R$50$, $+0.9$AP with R$101$ and $+0.7$ with Swin-L~\cite{liu2021swin}. Note that we only train \M for 72 epochs with Swin-L. On panoptic segmentation, our method achieves $+0.8$PQ with R$50$ and $+0.3$PQ with R$101$. Notably, our method is only a pluggable training method that will not affect the model architecture or inference speed. In addition, it only introduces negligible parameters (class embeddings) during training.
\subsection{Speed up training}

In Table~\ref{tab:speedup}, we show our \M can significantly speed up training on ADE20k and COCO. We do not report the results on Cityscapes because this dataset is relatively small and only needs  $90$k steps to converge in Mask2Former, which is not necessary for additional speedup. 
The results show that our method trained with $36$ epochs achieved comparable results with Mask2Former trained with $50$ epochs with both R$50$ and R$101$ as the backbone on both panoptic and instance segmentation. Moreover, our method trained with half number of training steps ($80$K steps) achieves comparable results with Mask2Former trained with $160$K steps. Note that we use the same batch size as Mask2Former. We also report the clock time. The results show that our method reduces the actual training time by large. For a fair comparison, we use the same number of GPUs to run \M and Mask2Former. Because there may exist differences in speed for different machines, we run our method and Mask2Former on the same machine to test the running time.

\subsection{Ablation Study}

We conduct an ablation study on the COCO instance segmentation task with R$50$ as backbone.

\noindent\textbf{Effectiveness of each component:} We show the effectiveness of each version of our method trained for $12$ epochs and $50$ epochs in Table~\ref{tab:component}, where we show the result of raw Mask2Former and the result after each version of our method is adopted. Without adding noises, our method achieves an improvement of $+0.9AP$. With multi-layer MP-training, our method achieves $+1.5$AP. We also find that label noise does not work when trained for $50$ epochs. Note that except for the last line of the table, other lines of our method apply MP training on the first $3$ layers.

\noindent\textbf{Apply MP training on multiple layers:} In Table~\ref{tab:num layers}, we show the effectiveness of applying GT masks on multiple layers. The results indicate that more layers of GT masks can lead to better performance. By adding noised GT masks in all layers, \M achieves $+0.8$AP.

\noindent\textbf{Add noises to GT masks:} In the last four rows of Table~\ref{tab:num layers}, we show the effectiveness of adding different types of noises. Note that without any noise, \M achieves $39.6$AP which is corresponding to the $4$th row in Table~\ref{tab:component}. Shifting noise and scaling noise do not work in our method. Point noise achieved $+0.3$ extra growth.

\noindent\textbf{Some hyper-parameter tuning:} Mask2Former adopts coarse-to-fine feature maps in decoder layers. We try to adopt the largest feature map ($1/8$ feature map) in all decoder layers. Table~\ref{tab:hyper param} shows that a coarse-to-fine manner is better than using the largest feature map in all layers for our \M. We also try to use matching to assign predictions to GT instances in the MP part. The result shows that matching can also improve the result compared with raw Mask2Former but the hard assignment is better.

\section{Conclusion}
In this paper, we have analyzed failure cases of Mask2Former and found a problem of inconsistent predictions between adjacent layers which potentially leads to inconstant optimization goals and low utilization of queries and therefore limits the performance of Mask2Former. In order to address the problem, we developed a new MP-Former, whose key components include multi-layer mask-piloted training with point noises and label-guided training. Our experimental results showed that the proposed training method effectively mitigates the inconsistent prediction problem. When evaluating on three challenging datasets (ADE20K, Citescapes, and MS COCO), our method achieved improvement by a large margin on all three image segmentation tasks (semantic, instance, and panoptic). In addition, our method only introduces little computation during training and no extra computation in inference time. Finally, our method can also significantly speed up training. Especially, when trained with half number of steps, our method exceeds Mask2Former on all three segmentation tasks on ADE20K with both R$50$ and Swin-L as the backbones. \\\textbf{Limitations:} We evaluated \M on Mask2Former, but our method may be useful to some other models such as MaskFormer and HTC. We will take it as our future work. 

{\small
\bibliographystyle{ieee_fullname}
\bibliography{11_references}

\begin{thebibliography}{10}\itemsep=-1pt

\bibitem{carion2020end}
Nicolas Carion, Francisco Massa, Gabriel Synnaeve, Nicolas Usunier, Alexander
  Kirillov, and Sergey Zagoruyko.
\newblock End-to-end object detection with transformers.
\newblock In {\em European conference on computer vision}, pages 213--229.
  Springer, 2020.

\bibitem{chen2019hybrid}
Kai Chen, Jiangmiao Pang, Jiaqi Wang, Yu Xiong, Xiaoxiao Li, Shuyang Sun,
  Wansen Feng, Ziwei Liu, Jianping Shi, Wanli Ouyang, et~al.
\newblock Hybrid task cascade for instance segmentation.
\newblock In {\em Proceedings of the IEEE/CVF Conference on Computer Vision and
  Pattern Recognition}, pages 4974--4983, 2019.

\bibitem{chen2017deeplab}
Liang-Chieh Chen, George Papandreou, Iasonas Kokkinos, Kevin Murphy, and Alan~L
  Yuille.
\newblock Deeplab: Semantic image segmentation with deep convolutional nets,
  atrous convolution, and fully connected crfs.
\newblock {\em IEEE transactions on pattern analysis and machine intelligence},
  40(4):834--848, 2017.

\bibitem{chen2017rethinking}
Liang-Chieh Chen, George Papandreou, Florian Schroff, and Hartwig Adam.
\newblock Rethinking atrous convolution for semantic image segmentation.
\newblock {\em arXiv preprint arXiv:1706.05587}, 2017.

\bibitem{chen2020scaling}
Liang-Chieh Chen, Huiyu Wang, and Siyuan Qiao.
\newblock Scaling wide residual networks for panoptic segmentation.
\newblock {\em arXiv preprint arXiv:2011.11675}, 2020.

\bibitem{cheng2020panoptic}
Bowen Cheng, Maxwell~D Collins, Yukun Zhu, Ting Liu, Thomas~S Huang, Hartwig
  Adam, and Liang-Chieh Chen.
\newblock Panoptic-deeplab: A simple, strong, and fast baseline for bottom-up
  panoptic segmentation.
\newblock In {\em Proceedings of the IEEE/CVF conference on computer vision and
  pattern recognition}, pages 12475--12485, 2020.

\bibitem{cheng2021mask2former}
Bowen Cheng, Ishan Misra, Alexander~G. Schwing, Alexander Kirillov, and Rohit
  Girdhar.
\newblock Masked-attention mask transformer for universal image segmentation.
\newblock 2022.

\bibitem{cheng2021maskformer}
Bowen Cheng, Alexander~G. Schwing, and Alexander Kirillov.
\newblock Per-pixel classification is not all you need for semantic
  segmentation.
\newblock 2021.

\bibitem{Cordts2016Cityscapes}
Marius Cordts, Mohamed Omran, Sebastian Ramos, Timo Rehfeld, Markus Enzweiler,
  Rodrigo Benenson, Uwe Franke, Stefan Roth, and Bernt Schiele.
\newblock The cityscapes dataset for semantic urban scene understanding.
\newblock In {\em Proc. of the IEEE Conference on Computer Vision and Pattern
  Recognition (CVPR)}, 2016.

\bibitem{Cordts2015Cvprw}
Marius Cordts, Mohamed Omran, Sebastian Ramos, Timo Scharw{\"a}chter, Markus
  Enzweiler, Rodrigo Benenson, Uwe Franke, Stefan Roth, and Bernt Schiele.
\newblock The cityscapes dataset.
\newblock In {\em CVPR Workshop on The Future of Datasets in Vision}, 2015.

\bibitem{dosovitskiy2020image}
Alexey Dosovitskiy, Lucas Beyer, Alexander Kolesnikov, Dirk Weissenborn,
  Xiaohua Zhai, Thomas Unterthiner, Mostafa Dehghani, Matthias Minderer, Georg
  Heigold, Sylvain Gelly, et~al.
\newblock An image is worth 16x16 words: Transformers for image recognition at
  scale.
\newblock {\em arXiv preprint arXiv:2010.11929}, 2020.

\bibitem{du2021simple}
Xianzhi Du, Barret Zoph, Wei-Chih Hung, and Tsung-Yi Lin.
\newblock Simple training strategies and model scaling for object detection.
\newblock {\em arXiv preprint arXiv:2107.00057}, 2021.

\bibitem{fang2021instances}
Yuxin Fang, Shusheng Yang, Xinggang Wang, Yu Li, Chen Fang, Ying Shan, Bin
  Feng, and Wenyu Liu.
\newblock Instances as queries.
\newblock In {\em Proceedings of the IEEE/CVF International Conference on
  Computer Vision}, pages 6910--6919, 2021.

\bibitem{ghiasi2021simple}
Golnaz Ghiasi, Yin Cui, Aravind Srinivas, Rui Qian, Tsung-Yi Lin, Ekin~D Cubuk,
  Quoc~V Le, and Barret Zoph.
\newblock Simple copy-paste is a strong data augmentation method for instance
  segmentation.
\newblock In {\em Proceedings of the IEEE/CVF Conference on Computer Vision and
  Pattern Recognition}, pages 2918--2928, 2021.

\bibitem{he2017mask}
Kaiming He, Georgia Gkioxari, Piotr Doll{\'a}r, and Ross Girshick.
\newblock Mask r-cnn.
\newblock In {\em Proceedings of the IEEE international conference on computer
  vision}, pages 2961--2969, 2017.

\bibitem{he2015deep}
Kaiming {He}, Xiangyu {Zhang}, Shaoqing {Ren}, and Jian {Sun}.
\newblock Deep residual learning for image recognition.
\newblock In {\em 2016 IEEE Conference on Computer Vision and Pattern
  Recognition (CVPR)}, pages 770--778, 2016.

\bibitem{huang2021fapn}
Shihua Huang, Zhichao Lu, Ran Cheng, and Cheng He.
\newblock Fapn: Feature-aligned pyramid network for dense image prediction.
\newblock In {\em Proceedings of the IEEE/CVF International Conference on
  Computer Vision}, pages 864--873, 2021.

\bibitem{kirillov2019panoptic}
Alexander Kirillov, Kaiming He, Ross Girshick, Carsten Rother, and Piotr
  Doll{\'a}r.
\newblock Panoptic segmentation.
\newblock In {\em Proceedings of the IEEE/CVF Conference on Computer Vision and
  Pattern Recognition}, pages 9404--9413, 2019.

\bibitem{li2022dn}
Feng Li, Hao Zhang, Shilong Liu, Jian Guo, Lionel~M Ni, and Lei Zhang.
\newblock Dn-detr: Accelerate detr training by introducing query denoising.
\newblock {\em arXiv preprint arXiv:2203.01305}, 2022.

\bibitem{li2022mask}
Feng Li, Hao Zhang, Shilong Liu, Lei Zhang, Lionel~M Ni, Heung-Yeung Shum,
  et~al.
\newblock Mask dino: Towards a unified transformer-based framework for object
  detection and segmentation.
\newblock {\em arXiv preprint arXiv:2206.02777}, 2022.

\bibitem{li2021panoptic}
Zhiqi Li, Wenhai Wang, Enze Xie, Zhiding Yu, Anima Anandkumar, Jose~M Alvarez,
  Tong Lu, and Ping Luo.
\newblock Panoptic segformer.
\newblock {\em arXiv preprint arXiv:2109.03814}, 2021.

\bibitem{lin2015microsoft}
Tsung-Yi Lin, Michael Maire, Serge Belongie, James Hays, Pietro Perona, Deva
  Ramanan, Piotr Doll{\'a}r, and C~Lawrence Zitnick.
\newblock Microsoft coco: Common objects in context.
\newblock In {\em European conference on computer vision}, pages 740--755.
  Springer, 2014.

\bibitem{liu2022dab}
Shilong Liu, Feng Li, Hao Zhang, Xiao Yang, Xianbiao Qi, Hang Su, Jun Zhu, and
  Lei Zhang.
\newblock Dab-detr: Dynamic anchor boxes are better queries for detr.
\newblock {\em arXiv preprint arXiv:2201.12329}, 2022.

\bibitem{liu2021swin}
Ze Liu, Yutong Lin, Yue Cao, Han Hu, Yixuan Wei, Zheng Zhang, Stephen Lin, and
  Baining Guo.
\newblock Swin transformer: Hierarchical vision transformer using shifted
  windows.
\newblock In {\em Proceedings of the IEEE/CVF International Conference on
  Computer Vision}, pages 10012--10022, 2021.

\bibitem{long2015fully}
Jonathan Long, Evan Shelhamer, and Trevor Darrell.
\newblock Fully convolutional networks for semantic segmentation.
\newblock In {\em Proceedings of the IEEE conference on computer vision and
  pattern recognition}, pages 3431--3440, 2015.

\bibitem{meng2021conditional}
Depu Meng, Xiaokang Chen, Zejia Fan, Gang Zeng, Houqiang Li, Yuhui Yuan, Lei
  Sun, and Jingdong Wang.
\newblock Conditional detr for fast training convergence.
\newblock {\em arXiv preprint arXiv:2108.06152}, 2021.

\bibitem{ren2015faster}
Shaoqing Ren, Kaiming He, Ross Girshick, and Jian Sun.
\newblock Faster r-cnn: Towards real-time object detection with region proposal
  networks.
\newblock {\em Advances in neural information processing systems}, 28, 2015.

\bibitem{ronneberger2015u}
Olaf Ronneberger, Philipp Fischer, and Thomas Brox.
\newblock U-net: Convolutional networks for biomedical image segmentation.
\newblock In {\em International Conference on Medical image computing and
  computer-assisted intervention}, pages 234--241. Springer, 2015.

\bibitem{vaswani2017attention}
Ashish Vaswani, Noam Shazeer, Niki Parmar, Jakob Uszkoreit, Llion Jones,
  Aidan~N Gomez, {\L}ukasz Kaiser, and Illia Polosukhin.
\newblock Attention is all you need.
\newblock In {\em Advances in neural information processing systems}, pages
  5998--6008, 2017.

\bibitem{wang2021max}
Huiyu Wang, Yukun Zhu, Hartwig Adam, Alan Yuille, and Liang-Chieh Chen.
\newblock Max-deeplab: End-to-end panoptic segmentation with mask transformers.
\newblock In {\em Proceedings of the IEEE/CVF Conference on Computer Vision and
  Pattern Recognition}, pages 5463--5474, 2021.

\bibitem{wang2021anchor}
Yingming {Wang}, Xiangyu {Zhang}, Tong {Yang}, and Jian {Sun}.
\newblock Anchor detr: Query design for transformer-based detector.
\newblock {\em arXiv preprint arXiv:2109.07107}, 2021.

\bibitem{xie2021segformer}
Enze Xie, Wenhai Wang, Zhiding Yu, Anima Anandkumar, Jose~M Alvarez, and Ping
  Luo.
\newblock Segformer: Simple and efficient design for semantic segmentation with
  transformers.
\newblock {\em Advances in Neural Information Processing Systems}, 34, 2021.

\bibitem{zhang2022dino}
Hao Zhang, Feng Li, Shilong Liu, Lei Zhang, Hang Su, Jun Zhu, Lionel~M Ni, and
  Heung-Yeung Shum.
\newblock Dino: Detr with improved denoising anchor boxes for end-to-end object
  detection.
\newblock {\em arXiv preprint arXiv:2203.03605}, 2022.

\bibitem{zhang2021k}
Wenwei Zhang, Jiangmiao Pang, Kai Chen, and Chen~Change Loy.
\newblock K-net: Towards unified image segmentation.
\newblock {\em Advances in Neural Information Processing Systems}, 34, 2021.

\bibitem{zheng2021rethinking}
Sixiao Zheng, Jiachen Lu, Hengshuang Zhao, Xiatian Zhu, Zekun Luo, Yabiao Wang,
  Yanwei Fu, Jianfeng Feng, Tao Xiang, Philip~HS Torr, et~al.
\newblock Rethinking semantic segmentation from a sequence-to-sequence
  perspective with transformers.
\newblock In {\em Proceedings of the IEEE/CVF conference on computer vision and
  pattern recognition}, pages 6881--6890, 2021.

\bibitem{zhou2019semantic}
Bolei Zhou, Hang Zhao, Xavier Puig, Tete Xiao, Sanja Fidler, Adela Barriuso,
  and Antonio Torralba.
\newblock Semantic understanding of scenes through the ade20k dataset.
\newblock {\em International Journal of Computer Vision}, 127(3):302--321,
  2019.

\bibitem{zhu2020deformable}
Xizhou {Zhu}, Weijie {Su}, Lewei {Lu}, Bin {Li}, Xiaogang {Wang}, and Jifeng
  {Dai}.
\newblock Deformable detr: Deformable transformers for end-to-end object
  detection.
\newblock In {\em ICLR 2021: The Ninth International Conference on Learning
  Representations}, 2021.

\end{thebibliography}
}

\newpage

\newpage
\appendix
\begin{table*}[t]
  \centering

  \tablestyle{4pt}{1.2}\scriptsize\begin{tabular}{l| lcc | x{20}x{20}x{20}x{20} |x{24}x{24}x{18}}
  method & backbone & query type & epochs & AP & AP$^\text{S}$ & AP$^\text{M}$ & AP$^\text{L}$  & \#params. & FLOPs & fps \\
  \shline
  Mask2Former~\cite{cheng2021mask2former}  & R50 & 100 queries & 12 &38.7 & 18.2 & 41.5& 59.8  & \phantom{0}44M & \phantom{0}226G & \phantom{0}9.7 \\
  Fixed matching scheme  & R50 & 100 queries & 12 &36.0 & 15.4 & 38.6& 57.2 & \phantom{0}44M & \phantom{0}226G & \phantom{0}9.7 \\
  Auxiliary loss  & R50 & 100 queries & 12 &38.2 & 17.7 & 41.1& 59.9  & \phantom{0}44M & \phantom{0}226G & \phantom{0}9.7 \\
  \end{tabular}
  \vspace{-1mm}

   \caption{The results of the naive solutions on COCO Instance segmentation. All methods in the table are trained for 12 epochs. Both solutions do not work.}


\label{tab:naive solutions}
\end{table*}
\section{Disscussion on noise types}
\label{supp:noise type}
\subsection{Why \M works even without noise?}
According to Table~\ref{tab:num layers}, MP training improves AP by 0.9 even without any noise, while in DN-DETR, the auxiliary denoising loss only works when the noise scale is $>0$. This difference is due to two reasons:
\begin{enumerate}
    \item DN-DETR adopt layer-by-layer refinement. It feeds GT boxes as the initial anchors.  Then, it predicts
offset $(\Delta x, \Delta y, \Delta w, \Delta h)$ in each layer and add it into
the last layer’s anchor. Without noises, the model can easily predict the GT by letting $(\Delta x, \Delta y, \Delta w, \Delta h)=(0, 0, 0, 0)$. Therefore, the model learns nothing when giving GT boxes. However, MP-Former feeds GT masks as attention masks which are not used to construct predictions. Therefore, even when GT masks are given, it is challenging to predict GT masks.
    \item Even though we do not give it noises, MP-Former has noises in itself. GT masks are interpolated into the resolution of the corresponding feature map before being fed into a decoder layer. The interpolation induces noises.
\end{enumerate}
\subsection{Why shifting and scaling noises are worse than point noises?}
In this section, we explain why shifting and scaling noise do not work for MP-Former while they are useful for DN-DETR. In short, the reason is that DN-DETR uses deformable attention to probe image features within a box while \M uses standard attention to probe image features according to a mask. The different approaches to query features lead to their need for different approaches to control the noise magnitude.

\begin{figure}[h]

    \centering

\centering

\centering
\subfloat[]{
\begin{minipage}[h]{0.3\linewidth}
\centering
\includegraphics[width=\linewidth]{resources/images/noise_type/point.png}
\end{minipage}%
}%
\centering
\subfloat[]{
\begin{minipage}[h]{0.3\linewidth}
\centering
\includegraphics[width=\linewidth]{resources/images/noise_type/shift.png}
\end{minipage}%
}%
\centering
\subfloat[]{
\begin{minipage}[h]{0.3\linewidth}
\centering
\includegraphics[width=\linewidth]{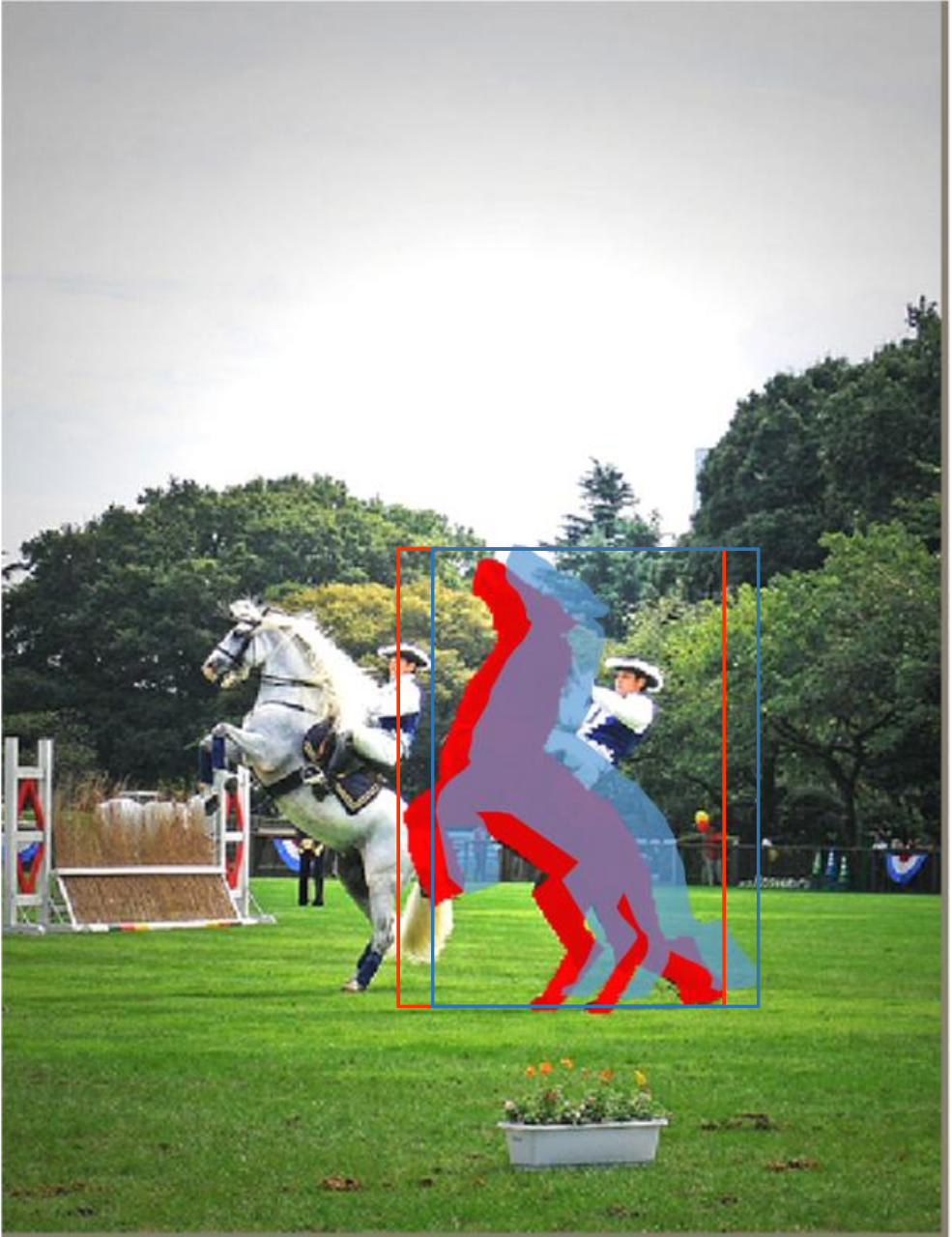}
\end{minipage}%
}%

\caption{This figure is to help illustrate the reason why point noises work in \M while shifting noises do not work. Red and blue regions are GT masks and noised masks, respectively. (a) Point noises on masks. (b) Shifting noises on masks (c) Shifting noises on boxes. Red and blue boxes are GT boxes and noised boxes, respectively.}
\label{fig:noise appendix}
\end{figure}

Following, we take shifting noise as an example to illustrate the reason in detail. Before explaining the detailed reason, we have to figure out why we add noise. Actually, the core idea of MP-Former is to give strong guidance to help the model learn ground-truth (GT) masks easier. The strongest guidance should be the GT masks, but it is easy for the model to learn GT masks when given GT masks. Therefore we add noises to make the prediction harder. Note that the noises should be small, otherwise the noised GT mask provides no guidance. The magnitude of the noises can be measured by the IoU of the noised masks (boxes) and GT masks (boxes). Small noises correspond to large IoU. Fig.~\ref{fig:noise appendix} shows examples of adding point and shifting noises on masks and adding shifting noise on boxes. Assuming we add point noises with $0.1$ as the noise ratio, the IoU will be as follows.
\begin{equation}
\begin{aligned}
    \frac{1-0.1}{1}\leq IoU &\leq \frac{1}{1+0.1}\\
    0.9 \leq IoU &\leq 0.91
\end{aligned}
\end{equation}
If we add a shifting noise of $0.1$ to the mask, the IoU is indeterminate. In the example shown in Fig.~\ref{fig:noise appendix} (b), the IoU is $< 0.7$. For shifting noises on masks, the IoU is dependent on not only the noise scale but also the shapes of the masks. 

However, when we add a shifting noise to a box, the IoU between a GT box and the noised box is determined. Assuming the noise scale is $\lambda$, we have $IoU=\frac{1-\lambda}{1+\lambda}$.
\section{Implementation details}
\label{supp:imp detail}
\noindent\textbf{The number of MP groups:}
Each image contains a different number of objects. Adopting a fixed number of MP groups is unreasonable. For images with few objects, the GPU memory will be wasted, and for those with many objects, memory may not be enough. To maximize the utilization of queries in the MP part, we adopt a dynamic number of MP groups. We adopt a fixed number of queries and the number of MP groups changes according to the number of objects in the image. Let $n_q, n_g, n_o$ denote the number of MP queries, MP groups, and objects. We have $n_g=\lfloor n_q/n_o\rfloor$.
\\\noindent\textbf{Self-attention mask: }
There are self-attention and cross-attention in Transformer decoder. Self-attention is applied among decoder queries. We find that queries in MP part bring key information about the GT masks. When doing self-attention, queries in the matching part will simply copy MP queries and learn nothing. Therefore, we adopt a self-attention mask to stop information leakage from the MP part to the matching part. This information leakage also exists among different MP groups, so we also adopt attention masks to stop inter-group information.
\\\noindent\textbf{Other details: }
Our models are trained on NVIDIA Tesla A100 GPUs with 40GB memory. All the models are trained with a total batch size 16, an initial learning rate $1.0\times 10^{-4}$, and a multi-step learning rate scheduler to drop twice by $0.1$ each. On ADE20k, our models with R50 and Swin-L backbone are trained on $2$ and $4$ GPUs, respectively. On Cityscapes, we train the model with $4$ and $8$ GPUs for our models with R50 and Swin-L backbone, respectively. On COCO2017, our models with R50 and R101 backbone are trained on $4$ and $8$ GPUs, respectively. All the hyper-parameters we do not mention are the same as Mask2Former.

\section{Some naive solutions for inconsistent mask predictions}
Before proposing our \M, We have tried some naive approaches to resolve inconsistent mask predictions. We introduce two of them and show the experimental results.
\begin{enumerate}
    \item \textbf{Fixed matching scheme:} A most direct approach to improve the $Util^i$ proposed in section~\ref{failure} is to use the same matching scheme for all layers. Therefore, we propose to only do bipartite matching for the last decoder layer and let the previous layers all use the same matching scheme.
    \item \textbf{Auxiliary loss:} A direct to improve mIoU-L$^i$ is to add a loss to encourage large mIoU-L$^i$. We formulate the loss as follows.
    \begin{equation}
        L=-\sum_{i=1}^9 \text{mIoU-L}^i
    \end{equation}
    Note that mIoU-L$^i$ is non-derivative. Therefore, we use mask loss of adjacent layers to approximate it as follows.
    \begin{equation}
        L=\sum_{i=1}^9 L_{mask}(M_i,M_{i-1})
    \end{equation}
    Where $L_{mask}(M_i,M_{i-1})$ denotes the mask loss with $M_i$ as prediction and $M_{i-1}$ as label.
\end{enumerate}

Table~\ref{tab:naive solutions} shows the performance of above mentioned naive solutions. We find fixed matching scheme does not work. When we observe the matched queries during training we find that only a part of queries is matched and optimized most of the time. So, we guess it is necessary for different layers to have different matching in the matching part. We also find directly enlarging the mIoU with an extra loss function does not work. Because the loss will encourage the predicted mask to follow the mask in the previous layer. Since the predicted masks in the first few layers are of low quality, the subsequent masks following the previous ones will also be bad. 

\section{Differences with DN-DETR:}
\begin{figure*}[t]

    \centering
\subfloat[]{
\begin{minipage}[h]{0.2\linewidth}
\centering
\includegraphics[width=1.in]{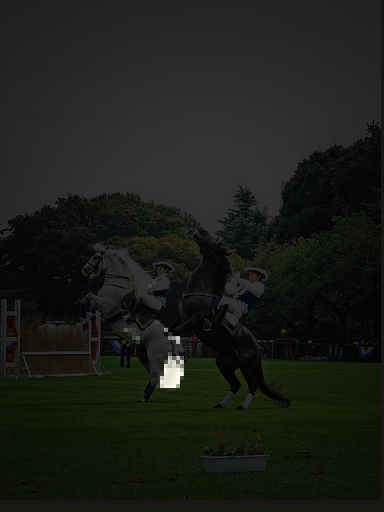}
\end{minipage}%
}%
\centering
\subfloat[]{
\begin{minipage}[h]{0.2\linewidth}
\centering
\includegraphics[width=1.in]{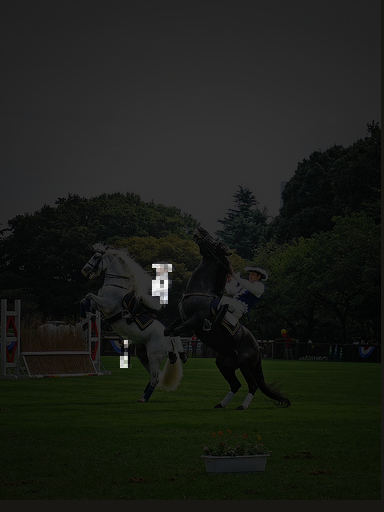}
\end{minipage}%
}%
\centering
\subfloat[]{
\begin{minipage}[h]{0.2\linewidth}
\centering
\includegraphics[width=1.in]{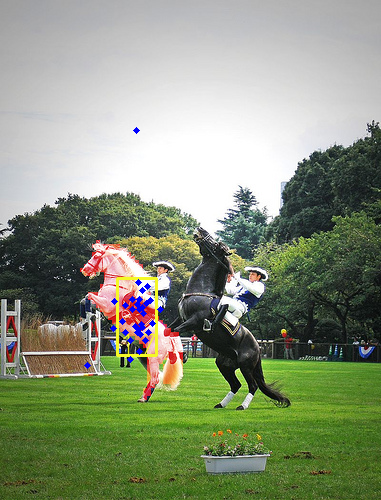}

\end{minipage}%
}%
\subfloat[]{
\begin{minipage}[h]{0.2\linewidth}
\centering
\includegraphics[width=1.in]{resources/images/tgt.png}

\end{minipage}%
}%
\centering
\subfloat[]{
\begin{minipage}[h]{0.2\linewidth}
\centering
\includegraphics[width=1.in]{resources/images/ly1_fail.png}

\end{minipage}%
}%





\caption{(a)(b) The attention map of the two ways to feed GT masks into decoder queries. (c) The blue points are the sampled locations in deformable attention and the yellow box is the predicted bounding box. Note that the corresponding GT object is the white horse.  (d)The red region is a given GT mask in first decoder layer which covers the man riding
the black horse.(e)The predicted mask of the first decoder layer which only covers the head of the man. The predicted mask will be used as the attention mask of the second layer and may mislead
the second layer.}
\label{fig:m2q}
\end{figure*}

\noindent\textbf{How to feed GT into decoder:}
 We had some initial attempts to adopt DN-DETR's way to feed GT through decoder queries. Since DN-DETR feeds GT boxes as positional queries to the Transformer decoder, we also tried to map noised GT masks into fixed-length position queries. The queries should contain part of the information about GT masks so that the model knows where to focus on to recover the GT masks. The first way we tried is to use the average of the image features within GT mask as the decoder queries of denoising part adding noises to the GT masks where we take features or directly add noises to the averaged vectors. However, it did not work. We find it difficult for the model to recover GT masks from averaged queries because taking averages may lead to losing much useful information about the GT masks. Another way we have tried is to map flattened $16\times 16$-resolution GT masks to position queries with an MLP. It still did not work. It seems difficult for the model to probe features within the GT masks as we want. Therefore, the main difference between our method and DN-DETR is that we feed GT masks as attention masks of the cross attention in Transformer decoder while DN-DETR feeds noised GT boxes as queries into the decoder. The root cause of the difference is the different attention mechanisms we are using. DN-DETR adopts deformable attention which samples positions around given reference points. While our method adopts standard attention which cannot fully use the position queries interpreted from GT masks. Fig.~\ref{fig:m2q} shows the attention map when we feed the mask as the query to probe features comparing with the sampled points by DN-DETR. It shows that it cannot find the right position to focus on when adopting the above-mentioned way.

\noindent\textbf{Multi-layer noises:}
In addition, we find adding mask guidance to multiple layers can further improve the performance as shown in Table~\ref{tab:num layers}. The more layers we add noised masks, the better the performance becomes. However, adding noised boxes to multiple layers does not work for DN-DETR. We visualize the output masks and boxes of the first decoder layer of our method and DN-DETR respectively in Fig~\ref{fig:m2q}(d)(e). we find that the output of DN-DETR is close to the GT boxes but some output masks of our model are far from GT masks. Therefore, giving mask guidance in multiple layer can further guide decoder layers to probe features. The root cause of this difference lies in the different ways of updating predictions. DN-DETR adopts layer-by-layer refinement. It predicts an offset $(\Delta x, \Delta y, \Delta w, \Delta h)$ in each layer and adds it into the last layer's prediction, which makes sure their box will not change much between layers. Therefore, when noised GT boxes are given in the first layer, the predictions of the following layers are usually close to GT boxes. Differently, Mask2Former adopts dot product to rebuild predictions in each layer. Therefore, it is more likely that their masks suffer great changes among layers.

\noindent\textbf{Noise types:}
Finally, we find the shifting and scaling way of adding noises in DN-DETR does not suit our method well. As shown in Table~\ref{tab:num layers}, point noises suit our method best. The potential reason and better ways to add noise will be left for future work.






\end{document}


\title{\paperTitle \\ Supplemental Material}
\author{\authorBlock}
\maketitle

\appendix
\label{sec:appendix}


{\small
\bibliographystyle{ieee_fullname}
\bibliography{11_references}
}